\documentclass[conference]{IEEEtran}
\IEEEoverridecommandlockouts
\usepackage{cite}
\usepackage{fancyhdr}
\usepackage{hyperref}
\usepackage{tikz}
\usepackage{array}
\usepackage{amsmath}
\usetikzlibrary{shapes.geometric, arrows}
\usetikzlibrary{shapes.geometric, arrows.meta, positioning, fit, backgrounds}
\usepackage{amsmath,amssymb,amsfonts}
\usepackage{algorithmic}
\usepackage{graphicx}
\usepackage{textcomp}
\usepackage{xcolor}
\usepackage{graphicx}
\usepackage{caption}
\usepackage{subcaption}
\def\BibTeX{{\rm B\kern-.05em{\sc i\kern-.025em b}\kern-.08em
    T\kern-.1667em\lower.7ex\hbox{E}\kern-.125emX}}
\begin{document}

\title{Estimating Vehicle Speed on Roadways Using RNNs and Transformers: A Video-based Approach}

\author{\IEEEauthorblockN{1\textsuperscript{st} Sai Krishna Reddy Mareddy}
\IEEEauthorblockA{\textit{Electrical and Computer Engineering Department} \\
\textit{University of North Carolina at Charlotte}\\
Charlotte, USA \\
smaredd2@uncc.edu}
\and
\IEEEauthorblockN{2\textsuperscript{nd} Dhanush Upplapati}
\IEEEauthorblockA{\textit{Electrical and Computer Engineering Department} \\
\textit{University of North Carolina at Charlotte}\\
Charlotte, USA \\
duppalap@uncc.edu}

\and 
\IEEEauthorblockN{3\textsuperscript{rd} Dhanush Kumar Antharam}
\IEEEauthorblockA{\textit{Electrical and Computer Engineering Department} \\
\textit{University of North Carolina at Charlotte}\\
Charlotte, USA \\
danthara@uncc.edu}
}

\maketitle

\begin{abstract}

This project explores the application of advanced machine learning models, specifically Long Short-Term Memory (LSTM), Gated Recurrent Units (GRU), and Transformers, to the task of vehicle speed estimation using video data. Traditional methods of speed estimation, such as radar and manual systems, are often constrained by high costs, limited coverage, and potential disruptions. In contrast, leveraging existing surveillance infrastructure and cutting-edge neural network architectures presents a non-intrusive, scalable solution. Our approach utilizes LSTM and GRU to effectively manage long-term dependencies within the temporal sequence of video frames, while Transformers are employed to harness their self-attention mechanisms, enabling the processing of entire sequences in parallel and focusing on the most informative segments of the data. This study demonstrates that both LSTM and GRU outperform basic Recurrent Neural Networks (RNNs) due to their advanced gating mechanisms. Furthermore, increasing the sequence length of input data consistently improves model accuracy, highlighting the importance of contextual information in dynamic environments. Transformers, in particular, show exceptional adaptability and robustness across varied sequence lengths and complexities, making them highly suitable for real-time applications in diverse traffic conditions. The findings suggest that integrating these sophisticated neural network models can significantly enhance the accuracy and reliability of automated speed detection systems, thus promising to revolutionize traffic management and road safety.

\end{abstract}


\section{Introduction}
In this project, we delve into the application of advanced neural network architectures, specifically Recurrent Neural Networks (RNNs) and Transformers, to estimate vehicle speeds using video surveillance data. Traditional speed estimation methods, such as radar and inductive loop detectors, while effective, are constrained by high costs, infrastructure demands, and limited geographic deployment. Our solution harnesses the sequential data processing power of RNNs, including their Long Short-Term Memory (LSTM) and Gated Recurrent Unit (GRU) variants, to capture dynamic temporal relationships in video data. These models are particularly adept at managing long-term dependencies, crucial for maintaining accuracy across longer sequences that depict vehicle movement.

Complementing the RNNs, we incorporate Transformers, which leverage self-attention mechanisms to process entire sequences simultaneously and focus selectively on the most relevant parts of the input data. This capability allows Transformers to excel in scenarios where contextual understanding from non-adjacent frames significantly enhances prediction accuracy. By embedding positional encodings, the model retains the sequential order of the data, a critical feature absent in traditional Transformer architectures but necessary for time-series analysis.

The integration of these sophisticated models aims to provide a robust, scalable, and cost-effective system for real-time vehicle speed detection. This system not only promises to improve traffic monitoring and safety but also introduces a method of leveraging existing camera infrastructure, thereby reducing the need for additional hardware investments. Through this innovative approach, we seek to revolutionize traffic management systems, offering enhanced accuracy and efficiency over conventional methods.

\section{DATA SET}

\subsection{VS13 Dataset}

The dataset utilized in this project, referred to as the VS13 dataset, is specifically designed for the task of vehicle speed estimation from video data. It comprises 400 high-definition video recordings capturing 13 distinct vehicle models, including Citroen C4 Picasso, Kia Sportage, and Mercedes AMG 550. Each vehicle in the dataset is recorded at various constant speeds ranging from 30 km/h to 105 km/h, with speeds maintained through onboard cruise control systems to ensure accuracy and consistency across different recordings.

Each video file in the dataset corresponds to a distinct vehicle and speed, labeled accordingly to facilitate easy identification and use in training and testing machine learning models. The dataset provides detailed temporal and spatial annotations for each frame within the videos. These annotations include bounding box coordinates that encase the vehicles, enabling precise tracking of their movement frame by frame.

The VS13 dataset is meticulously structured to support not only the evaluation of basic vehicle tracking and speed estimation algorithms but also the advanced capabilities of neural network architectures like RNNs and Transformers. The rich annotations and diverse conditions represented in the dataset make it an invaluable resource for developing robust models capable of accurately predicting vehicle speeds in various real-world scenarios. This comprehensive approach ensures that the models trained on this dataset are well-equipped to handle practical applications in traffic management and road safety enhancement.

\subsection{I-24 Multi-Camera 3D Dataset (I24-3D)}
The I24-3D dataset captures traffic scenes using 16-17 cameras, each recording at 4K resolution and 30 frames per second. This setup offers a densely covered 2000-foot segment of an interstate near Nashville, TN, allowing for high-fidelity tracking of vehicle movements across multiple lanes and varied traffic conditions.

Over 877,000 3D vehicle bounding boxes have been meticulously annotated by hand for 720 unique vehicles. These annotations provide critical spatial and temporal data, essential for training models to accurately detect and predict vehicle positions and speeds in three-dimensional space.

The dataset includes three different scenes, each representing sets of videos recorded simultaneously from various camera angles. This diversity allows models trained on this dataset to be robust against various lighting, weather, and traffic conditions.

The I24-3D dataset supports not only the development and testing of traffic monitoring systems but also advances research in multi-camera 3D tracking, a crucial aspect of modern traffic management technologies.

\section{APPROACH}
The project employs two advanced neural network architectures to tackle the problem of vehicle speed estimation from video data: Recurrent Neural Networks (RNNs), including their variants Long Short-Term Memory (LSTM) and Gated Recurrent Units (GRU), and Transformers. Each architecture offers unique benefits and is suited to particular aspects of the task.

\subsection{Feature Extraction}
In the project focused on vehicle speed estimation from video data, the feature extraction process is critical for capturing the dynamic properties of vehicles as they move across frames. The primary features extracted are based on differences in the coordinates of vehicle bounding boxes between consecutive frames. These include changes in the positions of the bounding box corners (x1, y1, x2, y2), which provide detailed information on the movement of each vehicle from one frame to the next.

Additionally, derived features such as the displacement in the center points of the bounding boxes and the changes in their dimensions (width and height) are calculated. These measurements are crucial as they directly relate to the vehicle's velocity and acceleration, which are fundamental metrics for speed estimation. 

In preprocessing, the dataset is initially cleansed and organized to ensure that each sequence of frames is sufficient to calculate the aforementioned features accurately. This involves selecting only those sequences where the number of frames meets a minimum threshold necessary for reliable feature calculation. The extracted features are then normalized to have zero mean and unit variance, enhancing model performance by providing a consistent scale for all input features. This normalization is essential for the effective training of neural networks, helping to ensure that no single feature dominates the learning process due to scale differences.

\subsection{RNN Approach}
\begin{itemize}
\item Getting Embeddings from Convolutional Layer : The first step in the processing pipeline involves a convolutional layer that transforms the input features—specifically, the bounding box dimensions of vehicles—into a higher-dimensional embedding space.
\item RNN Layer:  Depending on the specific requirements, various forms of RNNs can be utilized, such as basic RNNs, LSTMs, or GRUs. Each offers benefits in handling temporal dependencies, with LSTMs and GRUs providing enhanced capabilities for longer sequences due to their gated mechanisms.
\item Attention Mechanism:
Following the RNN layers, an attention mechanism is incorporated to refine the model's focus on crucial time steps within the sequence.
The attention module assesses and weights the importance of different frames based on their informational content regarding speed estimation. This selective focus allows the model to prioritize frames that contain significant movements or changes, enhancing the accuracy of the speed prediction.
\item Post attention modulation, the model employs another convolutional layer to process the attended features. This step is designed to further refine and enhance the representation of features that the model deems crucial.The refined features are then passed into a fully connected layer, which serves as the final step in the model architecture, outputting the predicted vehicle speed.

RNNs are inherently designed to handle sequences with their ability to maintain hidden states from previous inputs. This characteristic is particularly beneficial for analyzing video data where the temporal sequence and the continuity between frames are critical. In tasks like vehicle speed estimation, understanding how a vehicle’s position changes over time is crucial, and RNNs handle this temporal data naturally.

\end{itemize}
\tikzstyle{block} = [rectangle, draw, fill=white!20, text width=6em, text centered, rounded corners, minimum height=4em]
\tikzstyle{line} = [draw, -Latex]

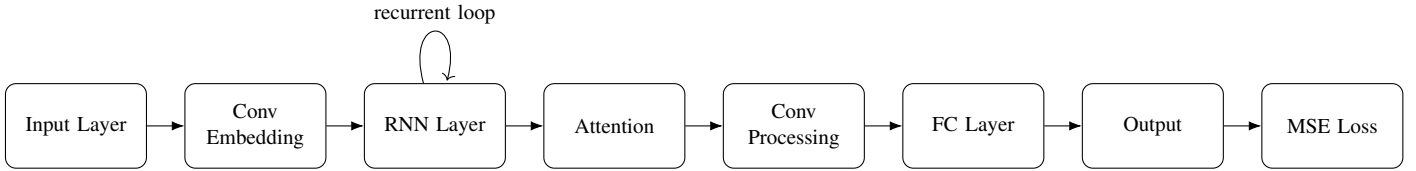
\begin{figure*}[!t]  
\centering
\begin{tikzpicture}[node distance=0.5cm, auto, scale=0.8, every node/.style={scale=0.8}]
    \node [block] (input) {Input Layer};
    \node [block, right=of input] (conv1) {Conv Embedding};
    \node [block, right=of conv1] (rnn) {RNN Layer};
    \node [block, right=of rnn] (attention) {Attention};
    \node [block, right=of attention] (conv2) {Conv Processing};
    \node [block, right=of conv2] (fc) {FC Layer};
    \node [block, right=of fc] (output) {Output};
    \node [block, right=of output] (loss) {MSE Loss};

    \path [line] (input) -- (conv1);
    \path [line] (conv1) -- (rnn);
    \path [line] (rnn) -- (attention);
    \path [line] (attention) -- (conv2);
    \path [line] (conv2) -- (fc);
    \path [line] (fc) -- (output);
    \path [line] (output) -- (loss);

    \path [line] (rnn) edge[loop above] node {recurrent loop} (rnn);
\end{tikzpicture}
\caption{Block diagram of the RNN-based vehicle speed estimation model}
\label{fig:blockdiagram}
\end{figure*}

\subsection{Transformer Based Approach}
The Transformer-based model architecture is constructed with several key components, each designed to process and enhance the video frame data effectively as shown in fig 2:
\begin{itemize}
 \item 1D Convolutional Embedding Layer: This layer serves as the initial entry point for input data, where bounding box dimensions or other spatial features from video frames are transformed into a higher-dimensional embedding space. This transformation is crucial as it allows the model to capture a richer representation of spatial dynamics before they are processed for temporal dependencies.
 \item Positional Encoding: Unlike traditional recurrent neural networks, Transformers do not inherently process data in sequence. To compensate for this, positional encodings are added to the embeddings to provide temporal context. This layer injects information about the relative or absolute position of the frames in the sequence, which is vital for maintaining the chronological order of events in the video data.
 \item Transformer Encoder: The core of the Transformer model is its encoder, which consists of multiple layers of self-attention and position-wise feedforward networks. Each layer in the encoder can attend to all positions in the previous layer simultaneously, making this model exceptionally good at modeling complex dependencies:
 Self-Attention: This mechanism allows the model to weigh the significance of different positions in the input data, depending on the task at hand. In the context of vehicle speed estimation, it enables the Transformer to focus more on frames that show significant movement or changes in the vehicle’s position.
Feedforward Networks: Each position outputs from the self-attention layer passes through a feedforward network that operates identically and independently on each position. This part of the model further transforms the data to help in the final prediction task.
\item Fully Connected Layers and Activation Functions: After processing through the Transformer encoder, the data passes through additional layers including fully connected layers and a ReLU activation function. These layers are used to map the encoded features to the final output space. A dropout layer is also included to prevent overfitting by randomly omitting subsets of features during training.

\item Output Layer: The final layer of the model outputs the predicted vehicle speeds. This layer typically involves a regression setup where the continuous speed values are predicted based on the learned features.

\end{itemize}
The Transformer model has shown exceptional performance in this project, particularly with longer sequences, which provide more contextual information. Its ability to process different parts of the input data simultaneously and its focus on relevant parts of the sequence make it highly effective for real-time vehicle speed estimation in diverse traffic conditions. The model's adaptability and robustness are crucial for deploying reliable speed estimation systems that need to operate across varying conditions and different types of traffic scenarios.

By leveraging this advanced Transformer architecture, the project not only enhances the accuracy and reliability of speed estimations but also pushes the boundaries of what can be achieved with AI in traffic management and vehicle monitoring systems.
\tikzstyle{block} = [rectangle, draw, fill=white!20, text width=5em, align=center, rounded corners, minimum height=3em]
\tikzstyle{line} = [draw, -Latex]
\tikzstyle{container} = [draw, rectangle, dashed, inner sep=2em]

\begin{figure*}[!t]  
\centering
\begin{tikzpicture}[node distance=0.5cm, auto, scale=0.9, every node/.style={scale=0.9}]
    \node [block] (input) {Input};
    \node [block, right=7mm of input] (embed) {Embedding};
    \node [block, right=7mm of embed] (posenc) {Pos Enc};
    \node [block, right=7mm of posenc] (encoder) {Encoder};
    \node [block, right=7mm of encoder] (fc1) {FC-1};
    \node [block, right=7mm of fc1] (relu) {ReLU};
    \node [block, right=7mm of relu] (dropout) {Dropout};
    \node [block, right=7mm of dropout] (output) {Output};
    
    \path [line] (input) -- (embed);
    \path [line] (embed) -- (posenc);
    \path [line] (posenc) -- (encoder);
    \path [line] (encoder) -- (fc1);
    \path [line] (fc1) -- (relu);
    \path [line] (relu) -- (dropout);
    \path [line] (dropout) -- (output);

    \begin{pgfonlayer}{background}
        \node[container, fit=(input) (output), label=above:Transformer Model] (container) {};
    \end{pgfonlayer}

\end{tikzpicture}
\caption{Block diagram of the Transformer-based vehicle speed estimation model}
\end{figure*}
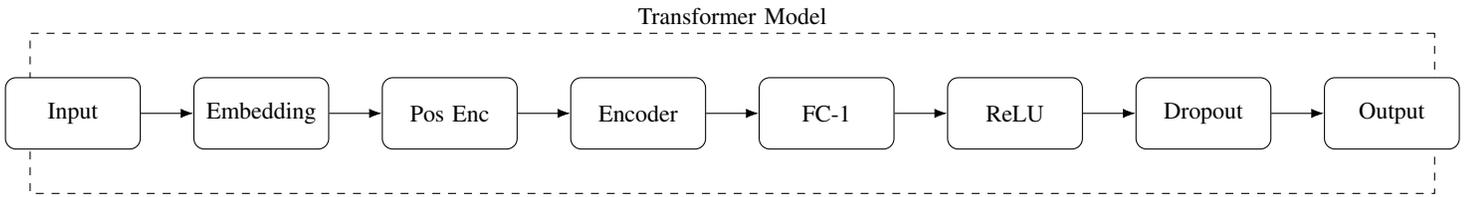

\subsection{Results}

The accuracy for each individual prediction is calculated using the formula:
\[
\text{Accuracy (\%)} = \left(1 - \frac{\left|\text{predicted} - \text{actual}\right|}{\left|\text{actual}\right|}\right) \times 100
\]
In addition to accuracy, the root mean squared error (RMSE) is calculated to provide another perspective on the model's performance. RMSE is useful for highlighting larger errors because it squares the residuals before averaging them, thus giving higher weight to larger errors.

$\text{RMSE} = \sqrt{\frac{1}{N} \sum_{i=1}^{N} (\text{predicted}_i -\text{actual}_i)^2}$

\begin{table}[htb]
\centering
\caption{Accuracy of Various Models on VS13 and I24\_3D Datasets}
\label{tab:model_accuracy}
\begin{tabular}{|c|m{2cm}|m{2.5cm}|}
\hline
\textbf{Model} & \textbf{VS13 Accuracy} & \textbf{I24\_3D Accuracy} \\ \hline
RNN            & 91.64      & 75.5         \\ \hline
LSTM           & 94.25      & 78.62           \\ \hline
GRU            & 92.66       & 77.88          \\ \hline
Transformers   & 90.90       & 74.36          \\ \hline
\end{tabular}
\end{table}

\begin{table}[htb]
\centering
\caption{RMSE of Various Models on VS13 and I24\_3D Datasets}
\label{tab:model_accuracy}
\begin{tabular}{|c|m{3cm}|m{2cm}|}
\hline
\textbf{Model} & \textbf{VS13 Dataset} & \textbf{I24\_3D Dataset} \\ \hline
RMSE GRU           & 5.08      & 11.7         \\ \hline
RMSE LSTM         & 3.96     & 10.99           \\ \hline
RMSE RNN         & 5.57       & 11.98          \\ \hline
RMSE Transformers   & 5.55       & 12.87          \\ \hline
\end{tabular}
\end{table}

\begin{figure*}[htb]
\centering
\begin{subfigure}{.5\textwidth}
  \centering
  \includegraphics[width=.9\linewidth]{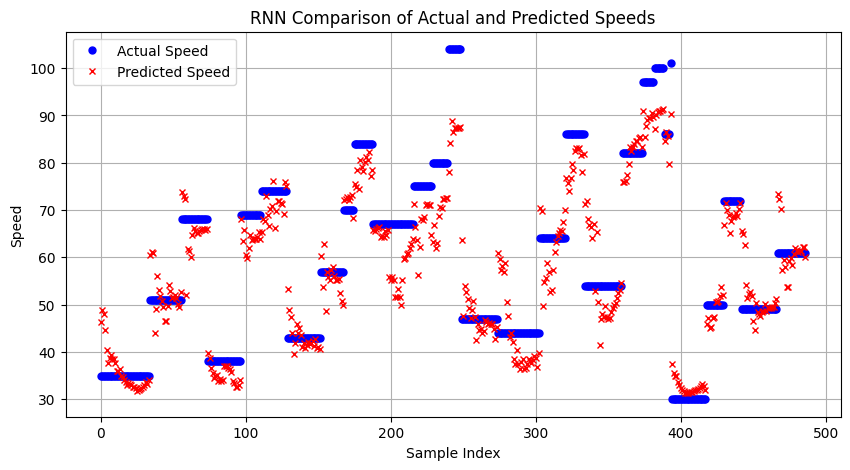}  
  \caption{RNN Model}
  \label{fig:sub-first}
\end{subfigure}%
\begin{subfigure}{.5\textwidth}
  \centering
  \includegraphics[width=.9\linewidth]{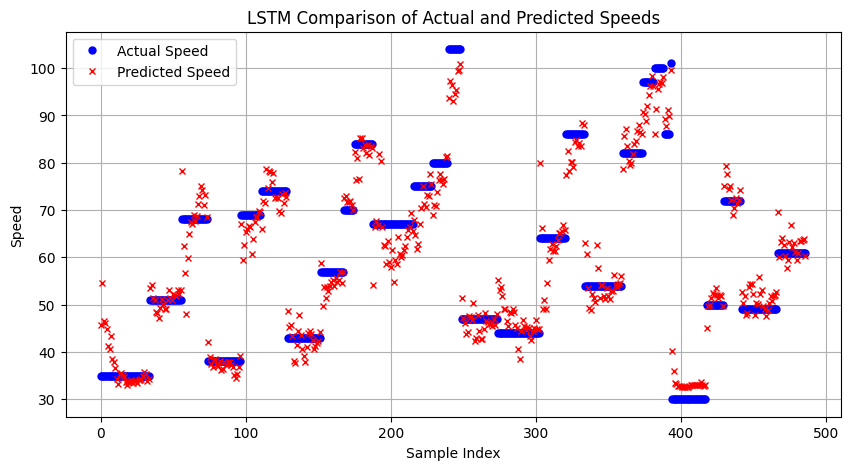}  
  \caption{LSTM Model}
  \label{fig:sub-second}
\end{subfigure}
\begin{subfigure}{.5\textwidth}
  \centering
  \includegraphics[width=.9\linewidth]{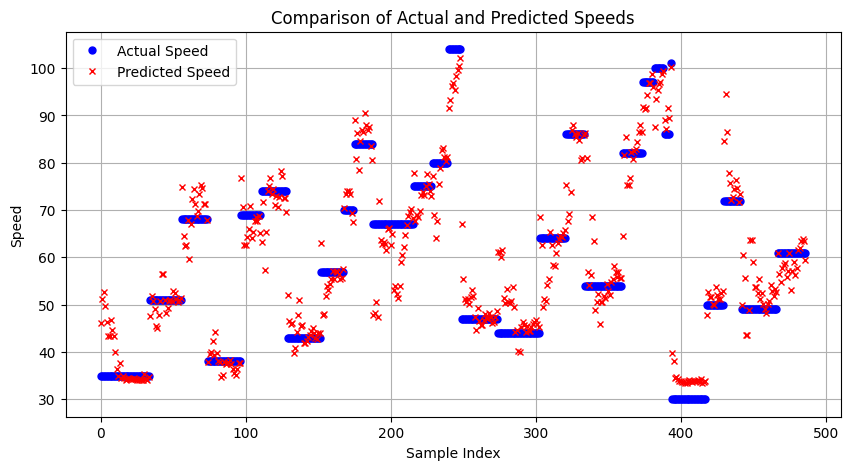}  
  \caption{GRU Model}
  \label{fig:sub-third}
\end{subfigure}%
\begin{subfigure}{.5\textwidth}
  \centering
  \includegraphics[width=.9\linewidth]{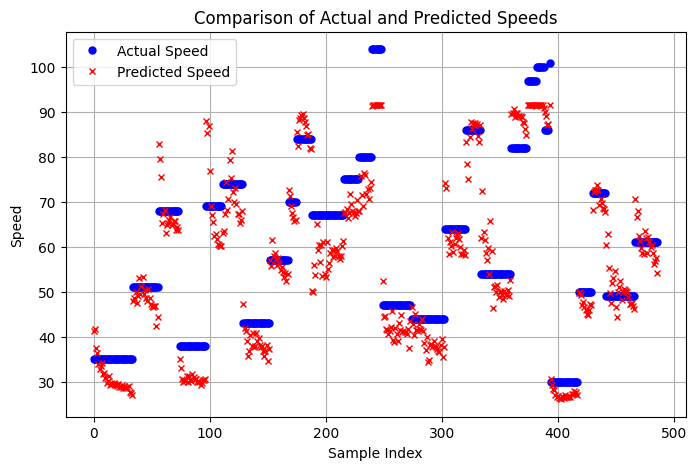}  
  \caption{Transformer Model}
  \label{fig:sub-fourth}
\end{subfigure}
\caption{Actual Speed vs Predicted Speed with VS13 Dataset }
\label{fig:fig}
\end{figure*}

\begin{figure*}[htb]
\centering
\begin{subfigure}{.5\textwidth}
  \centering
  \includegraphics[width=.9\linewidth]{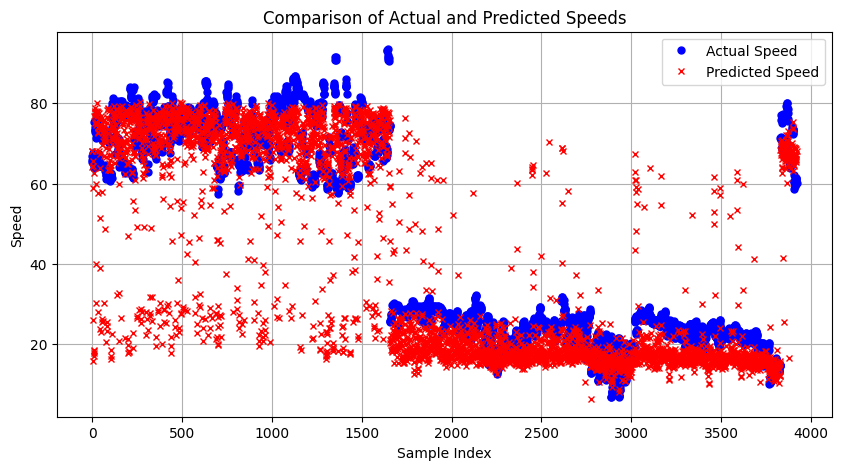}  
  \caption{RNN Model}
  \label{fig:sub-first}
\end{subfigure}%
\begin{subfigure}{.5\textwidth}
  \centering
  \includegraphics[width=.9\linewidth]{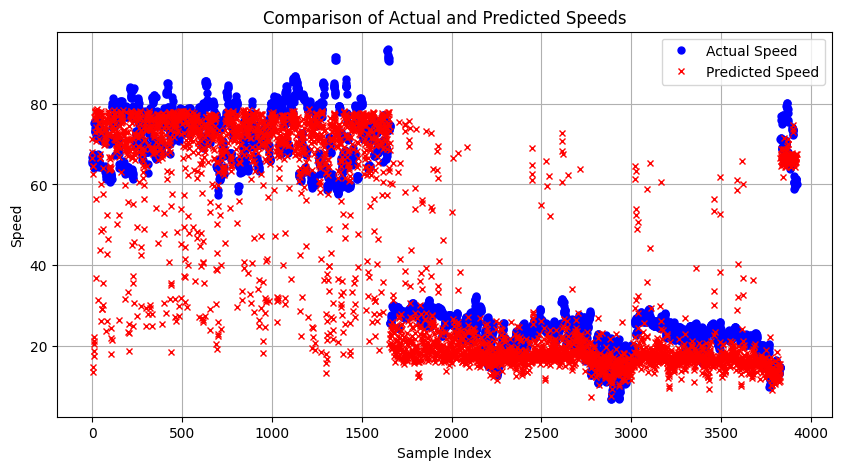}  
  \caption{LSTM Model}
  \label{fig:sub-second}
\end{subfigure}
\begin{subfigure}{.5\textwidth}
  \centering
  \includegraphics[width=.9\linewidth]{gru1.png}  
  \caption{GRU Model}
  \label{fig:sub-third}
\end{subfigure}%
\begin{subfigure}{.5\textwidth}
  \centering
  \includegraphics[width=.9\linewidth]{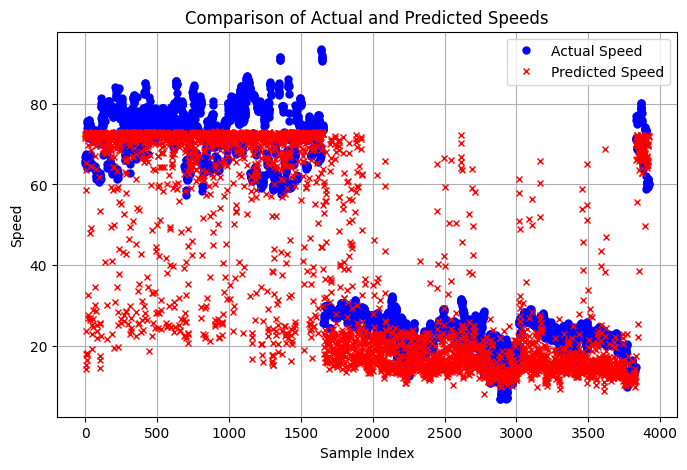}  
  \caption{Transformer Model}
  \label{fig:sub-fourth}
\end{subfigure}
\caption{Actual Speed vs Predicted Speed with I24 3D Dataset }
\label{fig:fig}
\end{figure*}
The exploration of various deep learning architectures for vehicle speed estimation from video data underscores significant insights into the capabilities of advanced recurrent models and Transformers. The analysis reveals that LSTM and GRU models, known for their ability to manage long-term dependencies, generally outperform basic RNNs. This advantage is particularly notable in tasks where the understanding of extended temporal sequences is crucial, highlighting the importance of advanced gating mechanisms present in these models.

Further, the study demonstrates that increasing the sequence length improves the performance of both RNNs and Transformers. By processing longer sequences, these models can capture more contextual information, thereby enhancing the accuracy and precision needed for both regression and classification tasks in vehicle speed estimation. This finding aligns with the intrinsic design of these models, which are optimized to synthesize and leverage extensive temporal data for more accurate predictions.

Transformers, distinguished by their self-attention mechanisms, exhibit exceptional performance, particularly with longer sequences. This capability allows them to dynamically focus on the most informative parts of the data, thereby enhancing prediction accuracy and robustness across varied and complex datasets. The adaptability of Transformers makes them especially suitable for real-time applications in diverse and unpredictable traffic conditions.

In conclusion, the comparative analysis of RNNs, LSTMs, GRUs, and Transformers in the context of vehicle speed estimation from video data not only highlights the specific strengths of each model type but also provides clear guidance for their application in traffic management systems. By choosing the appropriate model based on the specific requirements of the task at hand, such as the need for handling long-term dependencies or adapting to varied data complexities, developers can significantly enhance the reliability and accuracy of speed estimation systems. This contributes to safer and more efficient traffic management solutions, crucial for modern transportation infrastructures.

\subsection{Future Steps}
The project on vehicle speed estimation using advanced recurrent models and Transformer architectures has demonstrated promising results, highlighting the potential of deep learning techniques in traffic management systems. To further enhance the effectiveness and applicability of these models, the following future steps are recommended:
\begin{itemize}
\item Integration with Real-Time Systems:
While the models have shown high accuracy in controlled tests, integrating them into real-time traffic monitoring systems would be a crucial next step. This involves deploying the models on edge devices or cloud platforms capable of processing live video feeds from traffic cameras, testing their performance in real-time conditions, and ensuring they can operate efficiently under varying traffic scenarios.
\item Expansion of Dataset Diversity:
To improve the generalizability and robustness of the models, expanding the training datasets to include more diverse scenarios is essential. This includes different weather conditions, times of day, and traffic densities. More varied data will help the models better understand complex patterns and adapt to unforeseen situations.
\item  Exploration of Hybrid Models
Combining the strengths of LSTM, GRU, and Transformer models could lead to even more powerful systems for vehicle speed estimation. Hybrid models could leverage the sequential data processing of RNNs with the parallel processing capabilities of Transformers to enhance both accuracy and processing speed.
\item Enhanced Model Optimization
Further research into optimizing these models for better speed and lower computational costs is necessary, especially for deployment in environments with limited resources. Techniques such as model pruning, quantization, and knowledge distillation could be explored to reduce model size and computational demands without significantly compromising performance.
\item  Advanced Features Incorporation
Incorporating additional features such as vehicle type, size, and color recognition could enhance the accuracy of speed estimations. Advanced sensor data, such as LiDAR or radar, could also be integrated to provide more detailed inputs to the models.
\item Improved Error Handling and Reliability
Developing mechanisms to better handle errors and anomalies in input data, such as camera occlusions or faulty sensor data, would improve the reliability of the system. Implementing robust error handling strategies will ensure that the system remains accurate and dependable under less-than-ideal conditions.
\item User Feedback and Continuous Learning
Implementing a feedback system where the model can be continuously updated and improved based on user feedback and new data can help refine its predictions. Techniques such as online learning or reinforcement learning could be utilized to adaptively update the models based on their performance in the field.
\item Regulatory and Ethical Considerations
As these technologies might be used for law enforcement or may impact public safety, it's crucial to consider the regulatory and ethical implications. Ensuring transparency in how the models operate and making provisions for accountability in cases of errors are important steps.
\item Collaboration with Industry and Academia
Engaging in partnerships with academic institutions and industry stakeholders can provide additional resources, expertise, and real-world testing environments necessary for the advanced development and deployment of these technologies.

By pursuing these future steps, the project can not only enhance the technical capabilities of the models but also ensure they are practical, reliable, and ready for broader deployment in diverse and dynamic real-world environments.
\end{itemize}

\end{document}